\documentclass[letterpaper]{article} % DO NOT CHANGE THIS
\usepackage{aaai25}  % DO NOT CHANGE THIS
\nocopyright
\usepackage{times}  % DO NOT CHANGE THIS
\usepackage{helvet}  % DO NOT CHANGE THIS
\usepackage{courier}  % DO NOT CHANGE THIS
\usepackage[hyphens]{url}  % DO NOT CHANGE THIS
\usepackage{graphicx} % DO NOT CHANGE THIS
\usepackage{natbib}  % DO NOT CHANGE THIS AND DO NOT ADD ANY OPTIONS TO IT
\usepackage{caption} % DO NOT CHANGE THIS AND DO NOT ADD ANY OPTIONS TO IT
\usepackage{algorithm}
\usepackage{algorithmicx}
\usepackage{multirow}
\usepackage{subfigure}
\usepackage{booktabs}
\usepackage{arydshln}
\usepackage{amssymb}  
\usepackage{pifont}  
\usepackage{amsmath}
\usepackage{algpseudocode}
\usepackage{etoolbox}
\usepackage[most]{tcolorbox}
\newtcolorbox{mybox}[1][]{
breakable,
  arc=1mm,
  boxrule=1pt,
  colback=yellow!14,
  colframe=black!80,
  fonttitle=\bfseries,
  title=#1,
  left=1mm,
  right=1mm,
  top=1mm,
  bottom=1mm
}
\usepackage{xcolor}
\usepackage{newfloat}
\usepackage{listings}
\DeclareCaptionStyle{ruled}{labelfont=normalfont,labelsep=colon,strut=off} % DO NOT CHANGE THIS
\floatstyle{ruled}
\newfloat{listing}{tb}{lst}{}
\floatname{listing}{Listing}
\title{What Do You Want? User-centric Prompt Generation for Text-to-image Synthesis via Multi-turn Guidance}
\author {
    Yilun Liu\textsuperscript{\rm 1}\textsuperscript{$\ast$},
    Minggui He\textsuperscript{\rm 1}\thanks{Equal contribution.},
    Feiyu Yao\textsuperscript{\rm 1}\thanks{Corresponding author.},
        Yuhe Ji\textsuperscript{\rm 1},
    Shimin Tao\textsuperscript{\rm 1},
    Jingzhou Du\textsuperscript{\rm 1},
        Duan Li\textsuperscript{\rm 1},
    Jian Gao\textsuperscript{\rm 1},
    Li Zhang\textsuperscript{\rm 1},
        Hao Yang\textsuperscript{\rm 1},
    Boxing Chen\textsuperscript{\rm 2},
    Osamu Yoshie\textsuperscript{\rm 3}
}
\affiliations {
    \textsuperscript{\rm 1}Huawei, China\\
    \textsuperscript{\rm 2}Huawei Canada, Canada\\
    \textsuperscript{\rm 3}Waseda University, Japan\\
    \{liuyilun3, heminggui, yaofeiyu1, jiyuhe1, taoshimin, dujingzhou, li.liduan, gao.gaojian, izzie.zhangli, yanghao30, boxing.chen\}@huawei.com, yoshie@waseda.jp
}

\usepackage{bibentry}
\begin{document}

\maketitle

\begin{abstract}
The emergence of text-to-image synthesis (TIS) models has significantly influenced digital image creation by producing high-quality visuals from written descriptions. Yet these models heavily rely on the quality and specificity of textual prompts, posing a challenge for novice users who may not be familiar with TIS-model-preferred prompt writing. Existing solutions relieve this via automatic model-preferred prompt generation from user queries. However, this single-turn manner suffers from limited user-centricity in terms of result interpretability and user interactivity. To address these issues, we propose DialPrompt, a multi-turn dialogue-based TIS prompt generation model that emphasises user-centricity. DialPrompt is designed to follow a multi-turn guidance workflow, where in each round of dialogue the model queries user with their preferences on possible optimization dimensions before generating the final TIS prompt. To achieve this, we mined 15 essential dimensions for high-quality prompts from advanced users and curated a multi-turn dataset. Through training on this dataset, DialPrompt can improve interpretability by allowing users to understand the correlation between specific phrases and image attributes. Additionally, it enables greater user control and engagement in the prompt generation process, leading to more personalized and visually satisfying outputs. Experiments indicate that DialPrompt achieves a competitive result in the quality of synthesized images, outperforming existing prompt engineering approaches by 5.7\%. Furthermore, in our user evaluation, DialPrompt outperforms existing approaches by 46.5\% in user-centricity score and is rated 7.9/10 by 19 human reviewers.
\end{abstract}

\begin{links}
    \link{Code \& Datasets}{https://github.com/superboom/DialPrompt}
\end{links}

\section{Introduction}
\begin{figure}[!t]
 \centering  
 \subfigbottomskip=-2pt 
 \subfigcapskip=-2pt 
 \subfigure[Existing single-turn TIS prompt generation]{
  \includegraphics[width=0.96\linewidth]{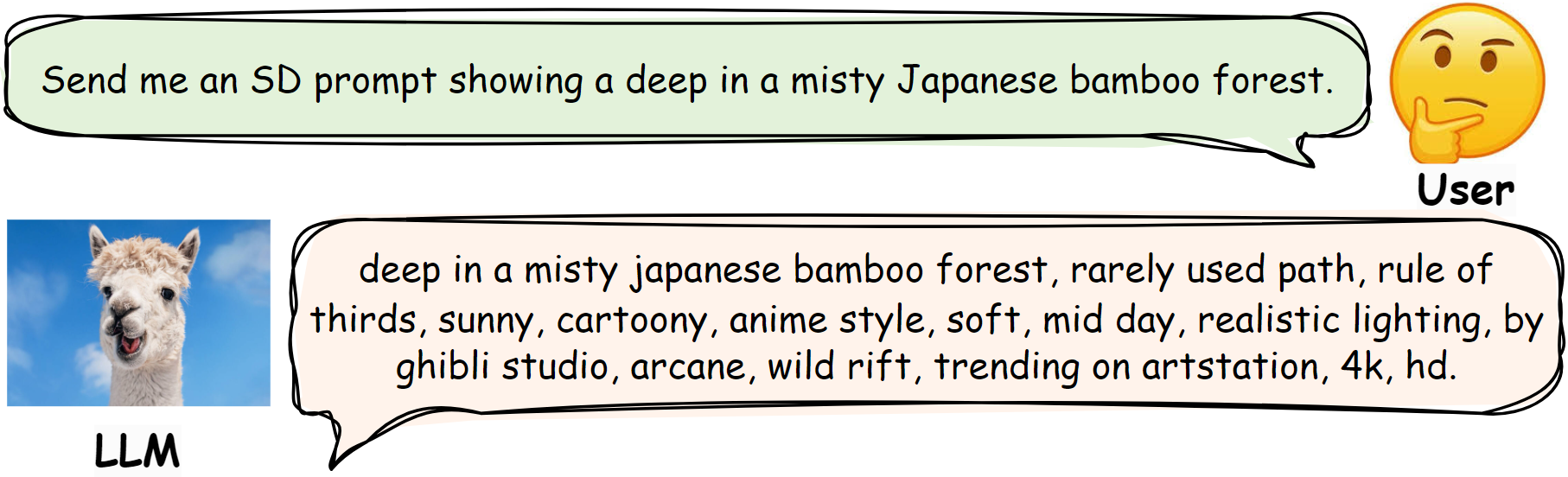}}
   \\
 \subfigure[Our DialPrompt]{
  \includegraphics[width=0.96\linewidth]{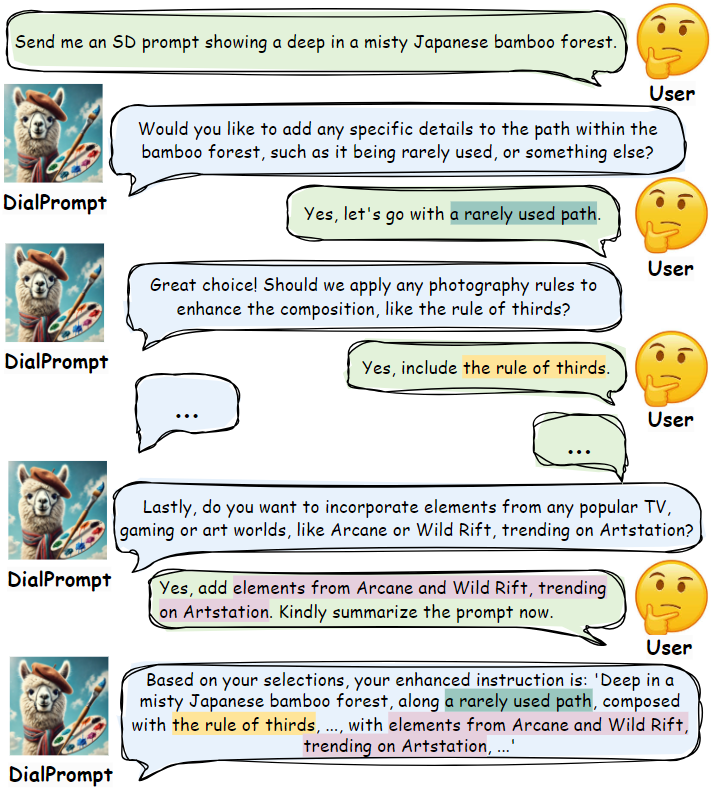}}
 \caption{Two user cases of TIS prompt generation with (a) single-turn style and (b) multi-turn guidance style.}
\label{fig_hook}
\end{figure}
The advent of text-to-image synthesis (TIS) models like Stable Diffusion (SD)~\cite{rombach2022high} has revolutionized the creation of digital images, enabling the generation of high-fidelity visuals from textual descriptions. However, as highlighted by recent studies~\cite{Large-scale_2023_ICIUI,Opal_2022_SUIST}, these models rely heavily on the quality of textual prompts provided by users. The specificity and relevance of these prompts may throw a significant impact on the fidelity and aesthetics of the generated images. Sometimes even adding some magical phrases in the prompts are key to a highly desirable image, such as ``soft'', ``by ghibli studio'' and ``arcane`` shown in Fig.~\ref{fig_hook}(a).

Thus, crafting the perfect model-preferred prompt for TIS models such as SD can be a challenging and nontrivial task for novice users who are not familiar with relevant keywords and prompt writing. While there has been research on manual principles of designing prompts to improve image quality~\cite{prompt-engineering_2022_CHI,best_prompts_2023_sigir}, an emerging trend is to assist novice users with automatic creation of model-preferred prompts from user-inputted descriptions~\cite{cao2023beautifulprompt,rosenman2024neuroprompts,hei2024user}. These approaches typically leverage Large Language Models (LLMs) to interpret user inputs and transform them into prompts that are more in line with the TIS model's preferences, thereby enhancing the aesthetic quality of the generated images.

However, we found existing single-turn-based approaches have several limitations in terms of user-centricity: 

\textbf{Firstly, interpretability remains a challenge.} Despite their ability to generate complex prompts, novice users often struggle to understand the significance of specific phrases within a prompt and how they correlate with the attributes of the generated image. For instance, as shown in Fig.~\ref{fig_hook}(a), after obtaining the complex prompt with a single-turn query, users may still be confused about the effectiveness of the added keywords, such as ``rule of thirds'', which actually controls the photography rule, and ``arcane'', which means adding elements from a popular television series. Furthermore, existing studies highlighted the challenge that users could face understanding barriers in why the model did not produce expected outputs, which hindered users' trust with models~\cite{zamfirescu2023johnny,weisz2023toward}.  

\textbf{Secondly, the existing methods suffer from a lack of interactivity.} Single-turn manners do not engage users in the prompt generation process, leading to outputs that may not align with the user's visual preferences. For example, in Fig.~\ref{fig_hook}(a), the user may desire a realistically styled image, but were provided with a prompt of a comic-style image. This is also observed in the study of Strobelt \emph{et al.}~\shortcite{strobelt2022interactive}, where they found that a prompt engineering tool should provide the user with the human-in-the-loop ability with rich feedback and user controlability to iteratively improve their prompt writing.

To address these shortcomings and enhance the user-centricity, we introduce DialPrompt, a dialogue-based TIS prompt generation model. DialPrompt seeks to improve upon the areas of interpretability and interactivity by conducting multiple rounds of queries to the user and gathering ample user preferences before generating the final prompt. To ensure user-centric experience, we studied 70k TIS prompts written by advanced users and mined 15 essential dimensions for crafting high-quality TIS prompts. Based on this finding, we curated a dataset containing 500+ multi-turn dialogues and trained DialPrompt. As shown in Fig.~\ref{fig_hook}(b), our multi-turn dialogue flow is designed to provide step-by-step guidance on possible directions of prompt optimization within the 15 dimensions, such as content, structure, art style, and atmosphere, thereby ensuring a better interpretability of the generated final prompt. Also, DialPrompt allows users to actively influence the outcome based on their specific visual preferences, thereby granting them greater control over the prompt generation process. Our contributions are summarized as follows:
\begin{itemize}
    \item We identified 15 essential dimensions for high-quality TIS prompts from advanced users, which can guide prompt engineering for TIS and lead to better visual effects of images, as indicated by DialPrompt's competitive generated image quality, outperforming existing TIS prompt generation models and general-purpose LLMs. 
    \item We proposed and validated a novel user-centric paradigm for TIS prompt generation that significantly enhances user experiences (with the ratings improved by 46.5\%) by allowing for more interpratable and personalized image creation processes.
    \item We open-sourced a high-quality dataset containing over 500 multi-turn dialogues for creating user-desired TIS prompts, facilitating future user-centric research.
\end{itemize}

\section{Related Work}
\subsection{Prompt Engineering in TIS}
Despite various architectures of TIS models proposed by researchers, such as autoregressive models~\cite{Zero-shot-text-to-image_2021_PMLP}, adversarial networks~\cite{sauer2023stylegan} and diffusion models~\cite{rombach2022high}, due to the relatively limited capacity of text encoders (such as the CLIP text encoder in SD~\cite{CLIP_2021_ICML}), they are still sensitive to quality of input prompts. The aim of prompt engineering in TIS is to organize prompts that achieve appealing visual effects of generated images. Since a TIS model was trained with a specific style of prompts, the philosophy of writing model-preferred prompts can be manually summarized, either by providing templates~\cite{best_prompts_2023_sigir} or magical keywords~\cite{taxonomy_prompt_modifiers_2022_arxiv,prompt-engineering_2022_CHI}. However, it still requires significant efforts for inexperienced users to choose suitable templates and master the keywords. To ease user's burden, by learning from vast exemplary prompts, various automatic TIS prompt generation models are proposed. Despite different training paradigms, they can be categorized into two classes in term of user experiences. The first is prefix-based, where user inputs a short prefix of their desired prompt and the model completes the prompt~\cite{rosenman2024neuroprompts,datta2023prompt,hao2024optimizing}. The second is instruct-based, where user inputs an instruction conveying their core ideas of creation and the model responds with a optimized prompt~\cite{manas2024improving,cao2023beautifulprompt,hei2024user}. 

Our work differs from existing approaches mainly in the user-machine interaction logic. Through a multi-turn dialogue, even novice users can be guided through in the optimization of prompt and fully express their preferences.   
\subsection{User-centric AI}
The aim of user-centric AI is to build explainable AI systems that users can understand, trust, and effectively manage~\cite{wang2019designing}. Various designing philosophies are proposed to achieve towards user-centric AI, including visual designing such as user interfaces~\cite{kim2023designing,feng2023promptmagician}, and procedure designing such as dialogue systems~\cite{cui2023chatedit,dong2024musechat}. Among them, the technique of reverse question answering (QA) is of particular interest~\cite{yin2019semi,yao2022deep}. In stead of answering user's questions, reverse QA systems ask user a series of questions in order to collecting preferences, thereby making the AI decision-making process more explainable and customized. Our work can be seen as a pioneering attempt to apply reverse QA into the field of TIS prompt generation to improve user-centricity.

\section{Methodology}

\subsection{Advanced User Observation}

The objective of DialPrompt is to enhance the interpretability and interactivity of TIS prompt generation by engaging users in a guided, step-by-step dialogue to capture their preferences. A critical prerequisite of this process involves identifying the key dimensions that define a high-quality TIS prompt. We achieved this goal by mining wisdom from advanced players of TIS models. Our initial dataset was sourced from lexica.art\footnote{\url{https://lexica.art/}}, a widely used platform for discovering SD images and prompts created and shared by experienced users. This platform can provide a comprehensive view of current best practices in TIS prompt engineering. We began with a publicly available dataset from Hugging Face\footnote{\url{https://huggingface.co/datasets/MadVoyager/stable_diffusion_instructional_dataset}}, which includes 70k advanced SD prompts collected from lexica.art, along with corresponding user instructions generated by LLMs. To enhance the dataset’s utility, we performed semantic clustering to remove redundant entries, ultimately refining it to a representative subset of around 5k pairs of user instructions and TIS prompts.

\begin{figure}[htbp]
 \centering  
    \includegraphics[width=0.9\linewidth]{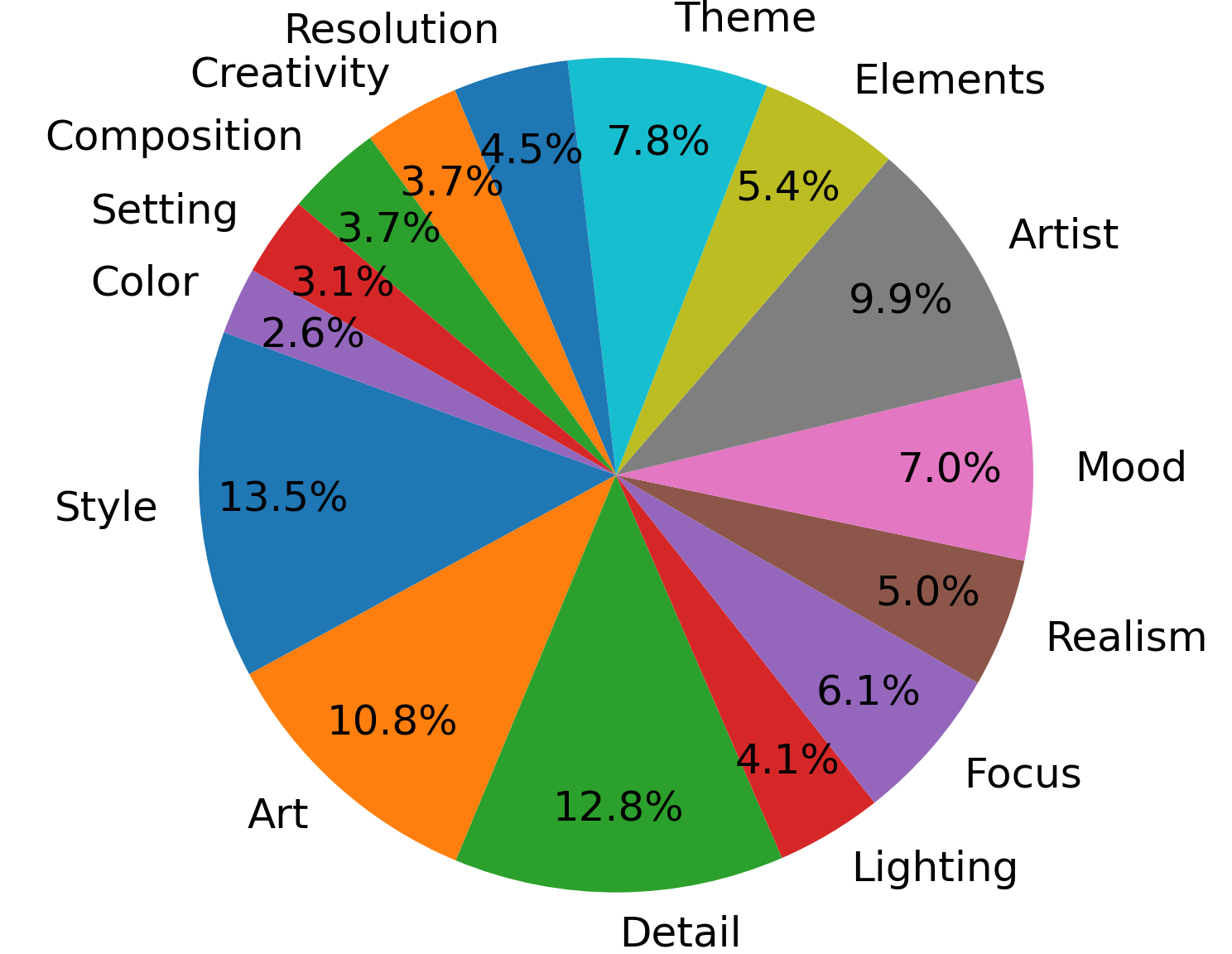}
 \caption{Occurrence distribution of 15 extracted dimensions in 5k advanced TIS prompts.}
\label{fig_table}
\end{figure}

We then actively work with a group of language experts, which are from the language service center of a top-tier corporation, and conducted a manual study on specific elements in the 5k TIS prompts. These prompts were evenly assigned to each language expert, who was asked to review the prompts and summarize key dimensions appeared in the prompts. After discussion with experts, we aggregated them into 4 major categories and 15 specific dimensions that are essential for crafting high-quality TIS prompts. The distribution of occurrences for these extracted dimensions within the 5k advanced TIS prompts is shown in the Fig.~\ref{fig_table}. Details of the extracted categories and dimensions are listed below:

\begin{figure*}[t!]
 \centering  
    \includegraphics[width=0.95\linewidth]{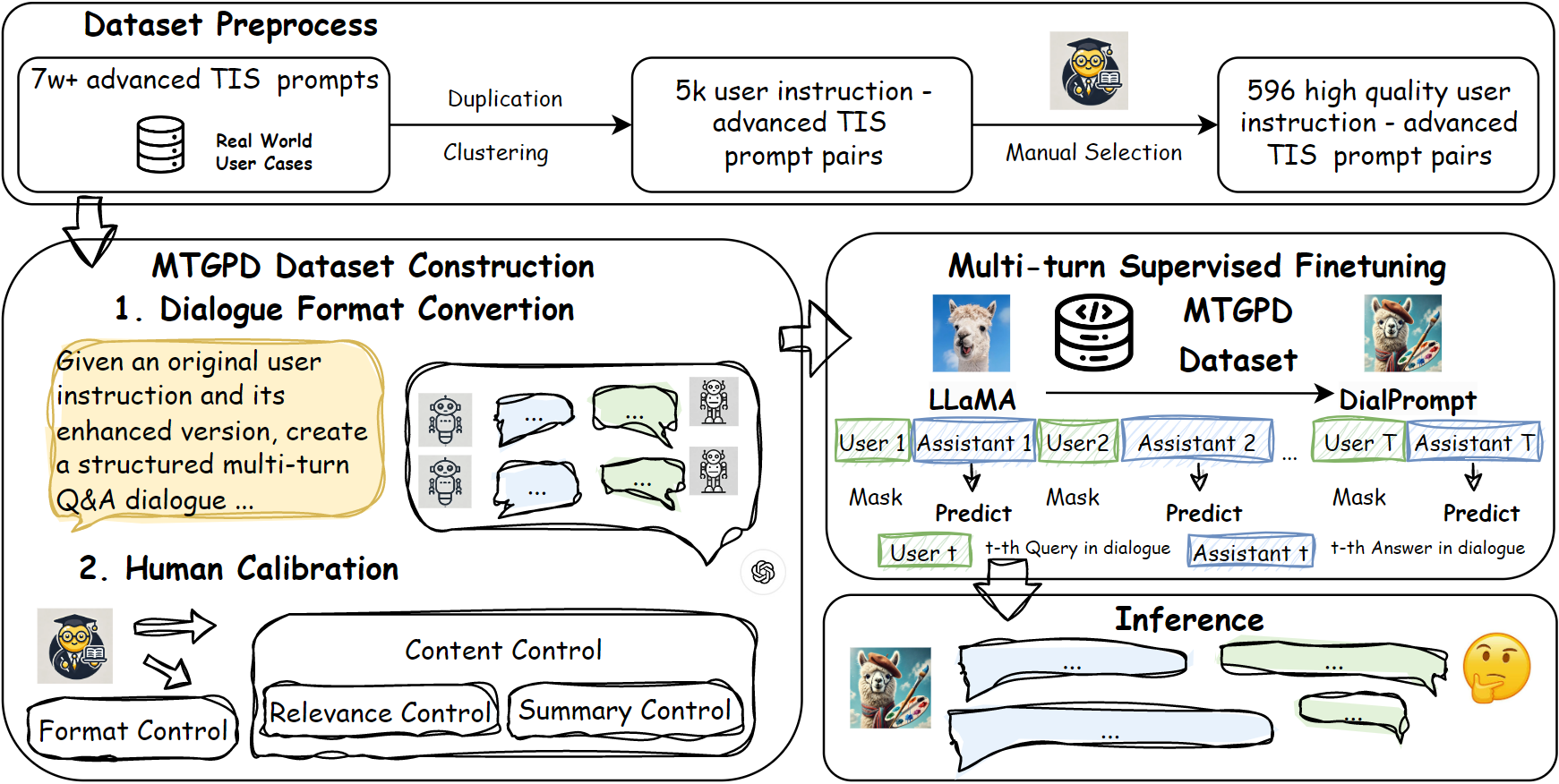}
 \caption{Illustration on the dataset construction, training and inference of DialPrompt.}
\label{fig1}
\end{figure*}

\begin{itemize}
    \item \textbf{Artistic Elements and Techniques}:
    This category encompasses the core components and methods of creating art, including \textbf{Style} (the visual appearance and artistic influences), \textbf{Art} (the various forms and media used), \textbf{Detail} (intricate aspects that enhance realism), and \textbf{Composition} (the arrangement of elements for visual balance).
    
    \item \textbf{Creative Expression}:
    This kind is focused on how artists convey ideas and emotions, including \textbf{Creativity} (innovation and uniqueness in art), \textbf{Theme} (the central subject guiding the narrative), and \textbf{Mood} (the emotional tone set by the artwork).
    
    \item \textbf{Visual Impact}:
    This group covers factors that influence the viewer’s perception, such as \textbf{Lighting} (use of light to affect atmosphere), \textbf{Focus} (primary points of interest), \textbf{Realism} (accuracy and lifelikeness), and \textbf{Color} (use of hues for emotional expression).
    
    \item \textbf{Context and Quality}:
    Background and quality of the artwork, including \textbf{Setting} (temporal and spatial context), \textbf{Resolution} (clarity and detail level), \textbf{Elements} (basic visual components like shapes and textures), and the \textbf{Artist} (whose style and skill shape the work).
\end{itemize}

\begin{algorithm}
\caption{GPT-4o's Workflow of Dialogue Construction from Instruction-Prompt Pairs}
\small 
\begin{algorithmic}[1]
\State \textbf{Input:} User Instruction Set $P_n = \{p_{n_1}, p_{n_2}, \dots, p_{n_N}\}$, Advanced Prompt Set $P_a = \{p_{a_1}, p_{a_2}, \dots, p_{a_N}\}$, Optimization Dimension Set $K = \{k_1, k_2, \dots, k_m\}$.
\For{each $(p_{n_i}, p_{a_i}) \in (P_n, P_a)$}
    \State \textbf{Step 1:} Compare dimension-specific differences \\ $\Delta_i = \text{diff}(p_{a_i}, p_{n_i})$.
    \State \textbf{Step 2:} Identify optimized dimensions\\ $K_i = \{k \in K \mid \Delta_{i,k} > \epsilon_k\}$.

    \For{each $k \in K_i$}
    \State \textbf{Step 3:} Compose a query $Q_{k}$ to user for dimension $k$ with optimization options.
    \EndFor

    \State \textbf{Step 4:} Convert $(p_{n_i}, p_{a_i})$ into dialogue format using composed queries $\{Q_{k}\}, k \in K$.
\EndFor
\State \textbf{Output:} Optimized dialogue format prompts in the form:
\State \hspace{\algorithmicindent} $\{(\text{User: } d_{n_1}, \text{System: } d_{a_1}), \dots, (\text{User: } d_{n_N}, \text{System: } d_{a_N})\}$.
\end{algorithmic}
\label{algorithm}
\end{algorithm}

These dimensions represent the key aspects that a high-quality TIS prompt should effectively address, thereby can guide through our construction of training dataset of DialPrompt. To mine a high-quality subset of TIS prompts, we established a filtering policy whereby any prompt that demonstrates enhancements in at least 5 of these dimensions is preserved. Language experts were then asked to manually label the dimensions in the 5k prompts and filter the unqualified ones, leading to a final selection of 596 high-quality advanced TIS prompt along with user instructions. These 596 high-quality data entries will serve as base for the generation of multi-turn dialogues for TIS prompt generation.

\subsection{Construction of MTGPD}

Based on the curated 596 high-quality pairs, we propose a multi-turn guidance prompt dataset (MTGPD). Each sample in the dataset is a representative dialogue between user and AI assistant, where the the AI assistant proactively asking users step-by-step questions to fulfill the initial user request and construct an final TIS prompt optimized in the 15 key dimensions as discussed above.

As shown in Fig.~\ref{fig1}, the construction of MTGPD is comprised of two primary components: Dialogue Format Conversion and Human Calibration. These components work synergistically to ensure that each dialogue in MTGPD is of the highest quality, both in terms of structure and content.

\subsubsection{Dialogue Format Conversion.}

The Dialogue Format Conversion process is designed to transform the 596 high-quality pairs of user instruction and advanced TIS prompt into a dialogue format. We use the assistance of GPT-4o~\cite{achiam2023gpt} to automation this dialogue generation task, with the workflow formally outlined in Algorithm \ref{algorithm}. The original prompt used for GPT-4o is in Appendix A.

As shown in Algorithm \ref{algorithm}, we utilize the previously extracted 15 key optimization dimensions as the foundation for the dialogue construction process. For each pair of user instruction and advanced TIS prompt, the user instruction is input as the start of a conversation. In each round, the user is presented with options corresponding to a specific optimization dimension within those in the corresponding advanced TIS prompt. The user selects their desired options, gradually refining the TIS prompts. After all dimensions existed in the advanced TIS prompts are discussed, the user terminates the conversation by ``Please summarize the prompt for me'' and the assistant summarizes the final TIS prompt.

\subsubsection{Human Calibration.}

To ensure the generated dialogue data meets high-quality standards, we implemented a Human Calibration process, which is critical for quality control. This process consists of three key steps: 

\textit{(1) Format Control:} To ensure that the generated dialogue data adheres strictly to a one-query-one-answer structure, avoiding instances where one party speaks multiple times consecutively, all generated dialogues undergo a rigorous format check. If any instances of consecutive speaking are detected, the dialogues are either corrected or excluded.

\textit{(2) Relevance Control:} To guarantee that the generated dialogue content is closely aligned with the topic, a semantic analysis is conducted on the generated dialogues to filter out content that does not contribute to the improvement of prompt quality, such as mutual compliments or expressions of thanks. Only content directly related to the optimization of the TIS prompt is retained.

\textit{(3) Summary Control:} To assure the completion, each constructed dialogue should include a final optimized prompt, marking the end of the conversation. If in the final round of dialogue, a summary of TIS prompt is not generated, the dialogue will be manually inspected and corrected.

\begin{table}[htbp]
\centering
\resizebox{0.96\linewidth}{!} {%
\begin{tabular}{|c|c|c|}
\hline

\multicolumn{3}{|c|}{\textbf{Statistics of Average tokens of dialogues}} \\ \hline
\multicolumn{2}{|c|}{Per user message} & 9.91 \\ \hline
\multicolumn{2}{|c|}{Per assistant message} & 28.26 \\ \hline
\multicolumn{2}{|c|}{\textbf{Statistics of Average round of dialogues}} & 6.16 \\ \hline
\multicolumn{3}{|c|}{\textbf{Prompt Optimize Dimensions}} \\ \hline
Style & Art & Detail \\ \hline
Creativity & Theme & Lighting \\ \hline
Focus & Realism & Setting \\ \hline
Mood & Resolution & Artist \\ \hline
Composition & Color & Elements \\ \hline
\multicolumn{2}{|c|}{\textbf{Average Number of Dimensions Per Dialogue}} & 6.99 \\ \hline

\end{tabular}
}
\caption{Statistics of MTGPD}
\label{tab:StatsOfMTGPD}
\end{table}

These processes ensure that the dataset is both structurally consistent and content-rich, thereby optimizing the performance and reliability of the model during training. The statistic of the final curated 596 dialogues in the MTGPD dataset is listed in Table~\ref{tab:StatsOfMTGPD}.

\subsection{Multi-Turn Supervised Fine-Tuning of DialPrompt}

For the training of DialPrompt, we developed a fine-tuning process specifically designed for our curated multi-turn dialogue dataset. This dataset provides rich conversational examples that allows the model to learn how to offer step-by-step guidance to users in optimizing TIS prompts across multiple key optimization dimensions. The fine-tuning process incorporates a \textit{multi-turn loss function}~\cite{zheng2024llamafactory}, as illustrated in Figure \ref{fig1}. During the training process, for a given dialogue sample from the MTGPD dataset, we apply a masking strategy to the user input. Dialprompt is then tasked with predicting only the assistant's responses. In the final loss computation, the total loss is calculated as the average of the cross-entropy losses for the predicted words in each assistant response throughout the conversation. This training strategy allows an efficient learning of assistant behaviors from the training sample by avoiding overly segmentation of multi-turn dialogues.

\section{Experiments}
\subsection{Experimental Setting}
\subsubsection{Implementation Details.}
The experiments are conducted on 8 NVIDIA A100 GPUs. In our implementation of DialPrompt, the MTGPD dataset is randomly split into a training set and a test set by a ratio of 9:1. DialPrompt is then trained on the training set for 10 epochs, with a learning rate of $1\times10^{-4}$, and a batch size of 16. The model is initialized from LLaMA3-8B-Instruct~\cite{dubey2024llama}. Stable Diffusion 3 Medium 
~\cite{esser2024scaling} is utilized as the TIS model in the main experiments. 
\subsubsection{User Preference Simulation.}
Given the multi-turn nature of DialPrompt, the generation of final TIS prompts require the other end of the dialogue, which is the participation of users. In mainstream multi-turn evaluation, the behavior of user end is fixed and irrelevant to AI responses, mostly asking pre-designed follow-up questions~\cite{zheng2024judging}. However, in the evaluation of DialPrompt, user needs to express their preferences on the suggestions and choices proposed by DialPrompt in each round of dialogue, which is unpredictable. Thereby, in addition to human evaluation, we also utilize GPT-4o-mini~\cite{achiam2023gpt} as an agent to enable an efficient user preference simulation. The prompt used in simulation for GPT-4o-mini is listed in Appendix B. To ensure the convergence of dialogues and avoid possible biases, the behavior of the agent is strictly prompted as following: (1) start the dialogue by querying with a user input in the test set; (2) respond with a random preference during the dialogue and (3) end the dialogue by asking for summarizing the prompt after a maximum number N of turns (We use N=5, approaching the average dialogue length in Table~\ref{tab:StatsOfMTGPD}).    
\subsubsection{Evaluation Dataset.}

In addition to the split test set from our MTGPD (which contains 60 samples and is denoted as MTGPD60), which is sourced from Lexica.art, another open-source TIS test set is also involved as an out-of-domain evaluation of DialPrompt. The out-of-domain test set, denoted as PP200, contains 200 prompts sampled from PartiPrompts~\cite{yu2022scaling}, which is designed to represent a wide range of topics, including different domains and features of language. For MTGPD60, we use the user instructions as the original user input to conduct TIS prompt generation. For PP200, we keep the prompts short by sampling only from the categories of Basic and Simple Detail, and directly use the short prompts as the user input prefix of prompt generation.
\subsection{Image Quality Evaluation}

After obtaining the generated TIS prompts using DialPrompt or other methods, we input them into Stable Diffusion-v3~\cite{esser2024scaling} to acquire the synthetic images. We then evaluate the quality of these images, which can reflect the overall quality of generated prompts. 
\subsubsection{Evaluation Metrics.} In the evaluation, We consider two dimensions of an image: \textit{fidelity} and \textit{aesthetic}. The dimension of \textit{fidelity} measures the degree to which the synthetic image reflects what the input prompt describes. As naive prompts often lead to deviated output images, an advanced prompt should steadily produce relevant images. Thus, we use CLIP Score~\cite{radford2021learning} as the metric of \textit{fidelity}, which measures the semantic consistency between the textual prompt and the produced image. For \textit{aesthetic}, we use Aesthetic Score~\cite{schuhmann2022laion}, which is a CLIP-based model trained on human aesthetic feedbacks to predict aesthetic score of images.
\subsubsection{Baselines.}
We consider two groups of baselines: (1) \textbf{TIS Prompt Models}. We compare DialPrompt with three prefix-based approaches: PromptGen\footnote{https://github.com/AUTOMATIC1111/stable-diffusion-webui-promptgen}, PromptExpansion~\cite{datta2023prompt} and MagicPrompt~\cite{cao2023beautifulprompt}, plus BeautifulPrompt~\cite{cao2023beautifulprompt}, which is a recent instruction-based model built upon LLMs. (2) \textbf{General-purpose LLMs}. Since most general-purpose proprietary LLMs nowadays are powerful in performing the task of prompt engineering~\cite{liu2024interpretable} and possess multimedia capabilities, we also include GPT-4o~\cite{achiam2023gpt}, GPT-3.5-turbo~\cite{ouyang2022training}, GPT-4-turbo~\cite{achiam2023gpt} and Claude-3.5-Sonnet~\cite{claude} in the evaluation, by directly instructing the LLMs to output an optimized prompt for Stable Diffusion.

\begin{table}[htbp]
\centering
\resizebox{0.9\linewidth}{!} {%
\begin{tabular}{lcccc}
\toprule
\multirow{2}{*}{\textbf{Method}} & \multicolumn{2}{c}{\begin{tabular}[c]{@{}c@{}}  \textbf{In-domain}\\ \multicolumn{1}{c}{\textbf{(MTGPD60)}} \end{tabular} }  & \multicolumn{2}{c}{\begin{tabular}[c]{@{}c@{}} \hspace{0.1em} \textbf{Out-of-domain}\\ \multicolumn{1}{c}{\textbf{(PE200)}} \end{tabular} } \\
\cmidrule(lr){2-3} \cmidrule(lr){4-5} 
 & \multicolumn{1}{c}{\textbf{CS}} & \multicolumn{1}{c}{\textbf{AS}} & \multicolumn{1}{c}{\textbf{CS}} & \multicolumn{1}{c}{\textbf{AS}} \\
 \midrule
\textbf{Original}                                 & 0.264                          & 5.913                               & 0.281                          & 5.542                               \\
\midrule
\multicolumn{2}{l}{\textbf{General-purpose LLMs}}            &                &                                &                                     \\
\midrule
GPT-3.5-turbo                                     & 0.302                          & 6.169                               & 0.284                          & 5.579                               \\
GPT-4-turbo                                       & 0.287                          & 6.311                               & 0.286                          & 5.569                               \\
GPT-4o                                            & \textbf{0.306}                 & 6.236                               & \textbf{0.296}                 & 5.728                               \\
Claude-3.5-Sonnet                                 & 0.280                           & 6.157                               & 0.288                          & 5.600                                 \\
\textbf{DialPrompt (ours)}                               & 0.287                          & \textbf{6.578}                      & 0.284                          & \textbf{6.263}                      \\
\midrule
\multicolumn{2}{l}{\textbf{TIS Prompt Models}}              &                &                                &                                     \\
\midrule
PromptGen                                         & 0.265                          & 5.925                               & 0.271                          & 5.376                               \\
PromptExpansion                                   & 0.267                          & 5.932                               & 0.276                          & 5.585                               \\
MagicPrompt                                       & 0.255                          & 6.000                               & 0.278                          & 5.531                               \\
BeautifulPrompt                                   & 0.263                          & 6.528                               & 0.255                          & 6.173                               \\
\textbf{DialPrompt (ours)}                               & \textbf{0.287}                 & \textbf{6.578}                      & \textbf{0.284}                 & \textbf{6.263}      \\
\bottomrule

\end{tabular}
}
\caption{Image quality scores on the in-domain test set and the out-of-domain test set. \textit{CS} and \textit{AS} stands for CLIP Score and Aesthetic Score. Best scores in each group are in \textit{bold}. \textit{Original} is images generated from original user inputs.}
\label{tab:as_result}

\end{table}
\subsubsection{Result.}

As shown in Table~\ref{tab:as_result}, DialPrompt not only significantly improves the image quality of original user inputs, but also outperforms that of existing TIS prompt models in all test cases. DialPrompt's advantage indicates its competitive TIS prompt optimization ability, which can lead to stable and visually-appealing images. For the comparison with general-purpose LLMs, despite curated with far less data and training pipelines, DialPrompt still outperforms existing LLMs in Aesthetic Score and achieves a comparable performance in CLIP Score. The enhancement in Aesthetic Score can be attributed to the comprehensive prompt optimization dimensions in the training data of DialPrompt, while the advantage in CLIP Score is a result of a more profound comprehension of user requests through multi-turn dialogues. Visualized cases are included in Appendix D.
\subsection{User Experience Evaluation}
In this section, we conduct a quantitative analysis on the user-centric experience provided by DialPrompt. To enable an efficient evaluation, both automatic (by GPT-4) and manual evaluation approach are utilized. 
\subsubsection{Evaluation Protocol.}

We define the following key dimensions of user experience to evaluate the extent of user-centricity an AI assistant demonstrates during interactions with users:
\begin{itemize}
    \item \textit{Clarity:} Language and layout clarity of AI's responses that allows users easily understanding generated content.
    \item \textit{Richness:} Richness of the AI recommended aesthetic elements in the dialogue where user can choose.
    \item \textit{Helpfulness:} Degree to which AI can understand user’s requirement and gives helpful guidance in dialogue.
\end{itemize}
Each evaluation dimension receives a score on a scale of 1 to 10, where a higher score indicates better performance.
\subsubsection{GPT-4 Evaluation.}

Following recent studies that utilize GPT-4 for evaluating LLM's capability~\cite{CoachLM}, we compose a prompt based on the above evaluation criterion to request evaluation results from GPT-4o (See Appendix C for the full prompt). In addition to scores from the three dimension, GPT-4o is also requested to output an overall score and a reason (to mitigate hallucination). The evaluation is comparison-based. GPT-4o is asked to compare user interaction processes from two AI assistants, given the dialogue records for every sample in the MTGPD60 test set. In the evaluation, we keep one of the assistant as the reference dialogues in MTGPD60 test set, which are human-calibrated, and the other assistant as the method to be tested. To mitigate biases, the final rating is the average of two tests, with the input order of the two assistants swapped. For the baselines, in addition to existing TIS prompt generation approaches, we also include general-purpose LLMs, since they also possess multi-medial and dialogue abilities. 

\begin{table}[htbp]
% \begin{center}
\centering
\resizebox{\linewidth}{!} {%
\begin{tabular}{lc@{\hskip 0.1in}c@{\hskip 0.1in}c@{\hskip 0.1in}c}
\toprule
\textbf{Method} & \textbf{Clarity} & \textbf{Richness} & \textbf{Helpfulness} & \textbf{Overall}\\
\midrule
\textbf{Reference Dialogue} & 8.67 & 8.48 & 8.67 & 8.65 \\
\midrule
PromptGen & 2.58 & 2.09 & 2.50 & 2.50 \\
PromptExpansion & 3.50 & 2.78 & 3.47 & 3.35 \\
MagicPrompt & 3.45 & 2.79 & 3.46 & 3.30 \\
BeautifulPrompt & 4.10 & 3.15 & 3.86 & 3.79 \\
Claude-3.5-Sonnet & 6.05 & 4.08 & 5.13 & 5.06 \\
GPT-3.5-turbo & 6.23 & 4.25 & 5.26 & 5.25 \\
GPT-4o & 6.17 & 4.13 & 5.14 & 5.15 \\
\hdashline
\noalign{\vskip 2pt}
\textbf{DialPrompt (ours)} & \textbf{7.81} & \textbf{7.57} & \textbf{7.72} & \textbf{7.69} \\
 \bottomrule
\end{tabular}
}
\caption{User-centricity score of different methods that generate TIS prompts for users. All reported scores for method $X$ are from $X$ \textit{v.s.} reference, except that the scores for reference dialogue are from DialPrompt \textit{v.s.} reference.}
\label{tab:GPT4-eva}
% \end{center}
\end{table}

The results are shown in Table~\ref{tab:GPT4-eva}. Due to their superior language abilities, general-purpose LLMs receive higher user-centricity scores than existing TIS prompt generation approaches. Nevertheless, DialPrompt outperforms both general-purpose LLMs and other prompt generation models in terms of Clarity, Richness and Helpfulness, indicating an advantage in achieving interpretable and interactive user experiences. Moreover, the overall rating of DialPrompt (7.69) reaches 88.9\% of the human-calibrated reference dialogues (8.65), which suggests DialPrompt's outstanding capabilities of user-centric TIS prompt generation.
\subsubsection{Human User Evaluation.}
Despite remarkable ratings given by LLMs, evaluations directly from human are irreplaceable. For this task, we recruited 19 well-educated volunteers with different backgrounds, which can be categorized into three groups. Group A contains seven professional visual designers from the design center of a top-tier corporation. They use TIS models such as SD to aid designing, and compose TIS prompts manually without tool assistance. Group B consists of six developers who are experienced users of TIS models. Around half of them have AI background and tried automatic prompt engineering. Group C are six amateur users who do not regularly use TIS models and are hardly exposed to prompt engineering technologies. Each reviewer is asked to independently conduct at least 10 fully completed dialogues with DialPrompt, acquiring optimized TIS prompts for different images that they desire. Then, they rate on Clarity, Richness and Helpfulness after their experiences with DialPrompt, according to the same criteria discussed in previous sections. We did not require image generation and the specific TIS model to use if they desire images. There is no overlap between volunteers and authors.

\begin{table}[htbp]

% \begin{center}
\centering
\resizebox{0.96\linewidth}{!} {%
\begin{tabular}{lccc}
\toprule
\textbf{Reviewers} & \textbf{Clarity} & \textbf{Richness} & \textbf{Helpfulness}\\
\midrule
Group A (Designer) & 8.71 & 6.71 & 7.71\\
Group B (Developer) & 8.33 & 7.33 & 6.50 \\
Group C (Amateur) & 7.58 & 7.83 & 7.50  \\
\hdashline
\noalign{\vskip 2pt}
\textbf{Average} & 8.21 & 7.29 & 7.24 \\
 \bottomrule
\end{tabular}
}
\caption{Average scores from human reviewers after at least 10 completed dialogues with DialPrompt.}
\label{tab:human_eval}
% \end{center}
\end{table}

The human evaluation result on DialPrompt is shown in Table~\ref{tab:human_eval}. The average scores in the three dimensions are close to GPT-4 evaluations in Table~\ref{tab:GPT4-eva}, indicating a steady performance of DialPrompt. Among the three groups, visual designers from Group A give the lowest average scores in Richness and the highest scores in Helpfulness, which suggests that the dialogue-based guidance from DialPrompt is an encouraging paradigm to optimize their workflow of art designing, but the richness of aesthetic elements is still not so satisfying from the angle of professional designers. In contrast, developers from Group B rate the two dimensions reversely. In stead of focusing on aesthetic elements, their behaviors during the dialogues are more flexible, and are not limited to linear dialogue flows, leading to a lower Helpfulness. For Group C, the amateur users give balanced scores for the three dimensions. This is in line with the original intention of DialPrompt, which is improving the experience of novice users in composing high-quality TIS prompts.    
\subsubsection{User Feedback.}

In addition to ratings, we also collect feedbacks from human reviewers. One of the reviewer commented: ``The dialogue style of DialPrompt is helpful. Based on the questions raised by DialPrompt, I can refine the specific scenarios I want to generate. I am glad to see the final image based on my own idea, and it is clear how the prompt is created.'' Another said: ``The recommended prompt elements are very professional, and can use as a reference in designing.'' And another commented: ``This tool can save the day for new users of SD. The AI offers easy-to-read suggestions on multiple angles that can optimize the prompt, making the image more visually pleasant.''

We also receive suggestions from reviewers. Several reviewers commented that the dialogue flow is designed to be too linear and users should be allowed to interact more with the AI, such as asking for further details and conducting open-domain discussions. Another frequent comment is to visualize the suggested prompt in each round of dialogue for a better understanding of the optimization process. These feedbacks and suggestions shed light on future directions of our work, such as incorporating reinforcement learning and multi-media training.
\subsection{Ablation Study}
\begin{table}[htbp]

% \begin{center}
\centering
\resizebox{0.82\linewidth}{!} {%
\begin{tabular}{lcc}
\toprule
\textbf{Method} & \textbf{CLIP Score} & \textbf{Aesthetic Score}\\
\midrule
Original & 0.264 & 5.913 \\
+Single-turn & 0.250 & 6.522 \\
+Multi-turn & \textbf{0.287} & \textbf{6.578}  \\
 \bottomrule
\end{tabular}
}
\caption{Image quality tested on MTGPD60 with different training styles.}
\label{tab:different_training_style}
% \end{center}
\end{table}
\subsubsection{Different Training Styles.} Instead of utilizing the full multi-turn dialogues in MTGPD as training data, we keep only the initial user query and the final optimized prompt in the dataset to train a single-turn TIS prompt model. As shown in Table~\ref{tab:different_training_style}, this single-turn model still significantly improves the Aesthetic Score of images from original user inputs, which suggests the effectiveness of the mining and cleaning process in the construction of MTGPD. Compared with single-turn, the multi-turn model improves largely in CLIP Score. Through multi-turn interaction with users, the TIS prompt generation process forms a step-by-step chain-of-thought~\cite{wei2022chain}, thereby decreasing hallucinations and increasing the stability of LLM performance.
\subsubsection{Different TIS models.}

\begin{table}[htbp]
\centering
\resizebox{0.76\linewidth}{!} {%
\begin{tabular}{lcccc}
\toprule
\multirow{2}{*}{\textbf{TIS Model}} & \multicolumn{2}{c}{\textbf{Original}}  & \multicolumn{2}{c}{\textbf{DialPrompt} } \\
\cmidrule(lr){2-3} \cmidrule(lr){4-5} 
 & \multicolumn{1}{c}{\textbf{CS}} & \multicolumn{1}{c}{\textbf{AS}} & \multicolumn{1}{c}{\textbf{CS}} & \multicolumn{1}{c}{\textbf{AS}} \\
 \midrule
LDM                                     & 0.278                          & 6.122                               & \textbf{0.296}                          &  \textbf{6.741}                               \\

SD-v1.5                                     & 0.267                          & 5.213                               & \textbf{0.273}                          &  \textbf{6.204}                               \\
SD-v2                                & 0.275                          & 6.134                               &  \textbf{0.294}                          &  \textbf{6.682}        \\

SDXL                                & 0.271                          & 6.119                               &  \textbf{0.295}                          &  \textbf{6.700}        \\

SD-v3                               & 0.264                          & 5.913                                &  \textbf{0.287}                 & \textbf{6.578}        \\
\bottomrule

\end{tabular}
}
\caption{Image quality of original user input and DialPrompt tested on MTGPD60 with different TIS models.}
\label{tab:different_TIS_model}
\end{table}

We test the same prompt on a series of different TIS models: LDM~\cite{rombach2022high}, SD-v1.5~\cite{rombach2022high}, SD-v2~\cite{rombach2022high}, SDXL~\cite{podellsdxl} and SD-v3~\cite{esser2024scaling}. As shown in Table~\ref{tab:different_TIS_model}, images generated from DialPrompt's optimized prompts continuously outperforms that from original user inputs, indicating a strong transferability of DialPrompt to different TIS models. 

\section{Conclusion}
In this paper, we seek to improve the user-centricity in TIS prompt engineering by proposing DiaPrompt, a novel dialogue-based TIS prompt generation model. DialPrompt not only shows advantages in improving the quality of synthetic images, but also provide a unique user experience through multi-turn guidance. Our user evaluation demonstrate that DialPrompt can not only assist novice users to easily optimize TIS prompt with their own ideas, but also aid professionals in their designing work through a comprehensive recommendation of aesthetic elements. Future work includes improving the flexibility of dialogues by incorporating reinforcement learning and increasing the prompt quality through multi-media training.

\bibliography{aaai25}

\newpage
\appendix
\section{Appendix A: Prompt Template for Dialogue Format Conversion via GPT-4}
\begin{mybox}
\verb|[System]|

Given an original user instruction and its enhanced version, create a structured multi-turn Q\&A dialogue that guides a user to refine their prompt for creating an aesthetically superior image. 
Here are notices: 

1. The dialogue should start with the user's original instruction.

2. This instruction is for Stable Diffusion to generate images.

3. Questions focus on providing enrich information for user to choose based on the enhanced description.

4. With user's chosen options , assistant output a final enriched instruction which is similar to enhanced description or keep enhanced description as final output.

5. The dialogue should be ended with a enhanced version prompt for Stable Diffusion. When you output the enhanced version prompt, please add \#\#\#[BEGIN OF PROMPT] before the prompt, and \#\#\#[END OF PROMPT] after the prompt like this: \#\#\#[BEGIN OF PROMPT] 'lofi biopunk portrait of Shrek as a Disney Princess, Pixar style, by Tristan Eaton, Stanley 'Artgerm' Lau, and Tom Bagshaw.' \#\#\#[END OF PROMPT]

6. Please Generate the multi-turn Q\&A dialogue with llama alpaca format of json file, the Role Name in each sample must be 'user' and 'assistant'.

7. In the last dialogue turn, you must add the summarize order for user content, like: 'user' : Artgerm and Greg Rutkowski. Please summarize the prompt for me now.

\verb|[User]|

Please Generate the multi-turn Q\&A dialogue with llama alpaca format of json file: \textcolor{red}{\{Input Instruction-Prompt Pair\}}
\end{mybox}

\section{Appendix B: Prompt Template for User Preference Simulation via GPT-4}
\begin{mybox}
\verb|[System]|

Assume you are a user who has a dialogue with a system which aims to enrich prompt for text to image generation, make suitable selection according to the option it provided without any biases, or ask it to combine all options based on your situation. your answer must be concise.

e.g. system: To create a more captivating image, would you like the portrait to be realistic or stylized? Your answer: Realistic, please. Or you can answer: A mix of both is ok.

\verb|[User]|

\textcolor{red}{\{Input Dialogue\}}
\end{mybox}

\section{Appendix C: Prompt Template for User Experience Evaluation via GPT-4}
\begin{mybox}
\verb|[The Start of Assistant 1’s Dialogue]|

\textcolor{red}{\{Input Dialogue 1\}}

\verb|[The End of Assistant 1’s Dialogue]|

\verb|[The Start of Assistant 2’s Dialogue]|

\textcolor{red}{\{Input Dialogue 2\}}

\verb|[The End of Assistant 2’s Dialogue]|

\verb|[System]|

We would like to request you to compare on the performance of two AI assistants in the displayed multi-turn dialogues with the user, trying to recommend and build a proper Stable Diffusion prompt for the user. Please rate the user-friendliness of the two AI assistants, considering the Clarity, Richness and Helpfulness of the whole dialogue. (1) Clarity: to which degree the layout and language of AI’s responses is organized and clear for users. (2) Richness: the richness of the AI recommended aesthetic elements that user can express preferences on in the dialogue. (3) Helpfulness: the degree to which the AI can understand user’s requirement and give step-by-step guidance in the dialogue. Each dimension receives a score on a scale of 1 to 10, where a higher score indicates better performance. And also output an overall score of 1 to 10. Please first output two lines indicating the scores for Assistant 1 and 2, with each line containing only four values indicating the scores for overall, clarity, richness and helpfulness, respectively. The four scores are separated by space. In the subsequent line, please provide a comprehensive explanation of your evaluation, avoiding any potential bias and ensuring that the order in which the dimensions were presented does not affect your judgment.
\end{mybox}

\section{Appendix D: Visualized Cases}
\begin{figure*}[tp]
 \centering  
 \subfigbottomskip=-2pt 
 \subfigcapskip=-2pt 
 \subfigure{
  \includegraphics[width=0.9\linewidth]{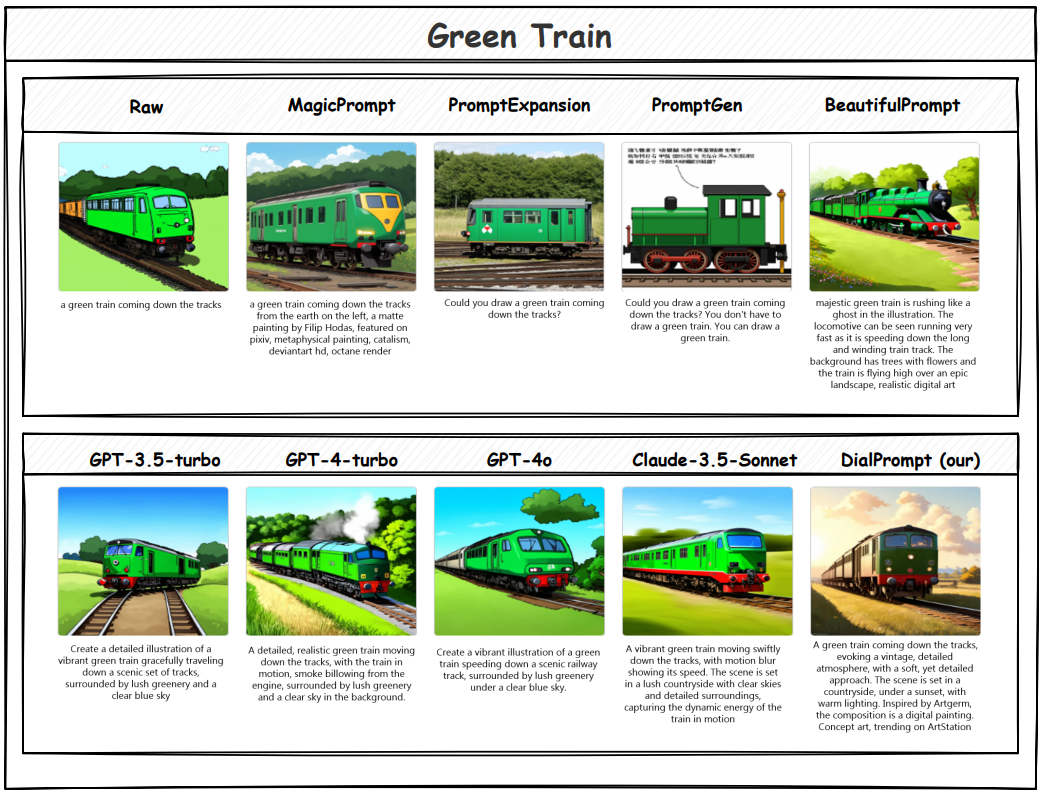}}
   \\
 \subfigure{
  \includegraphics[width=0.9\linewidth]{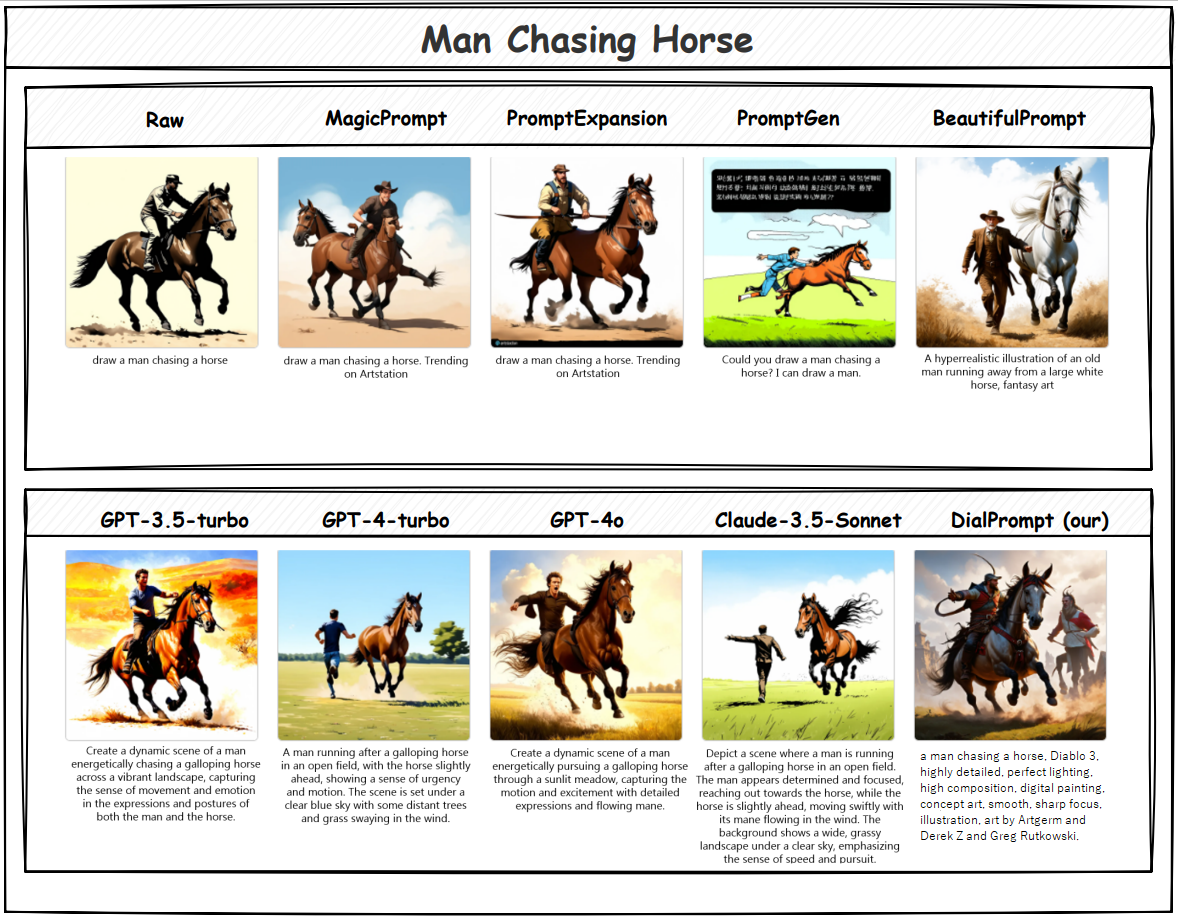}}

\end{figure*}
\begin{figure*}[tp]
 \centering  
 \subfigbottomskip=-2pt 
 \subfigcapskip=-2pt 
 \subfigure{
  \includegraphics[width=0.9\linewidth]{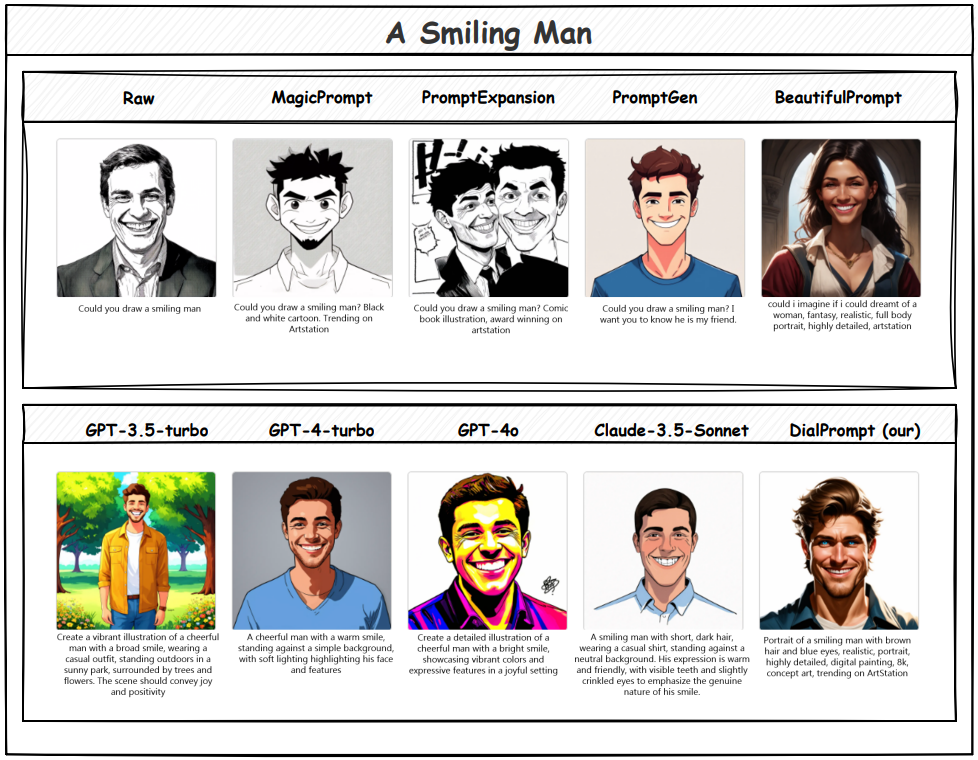}}
   \\
 \subfigure{
  \includegraphics[width=0.9\linewidth]{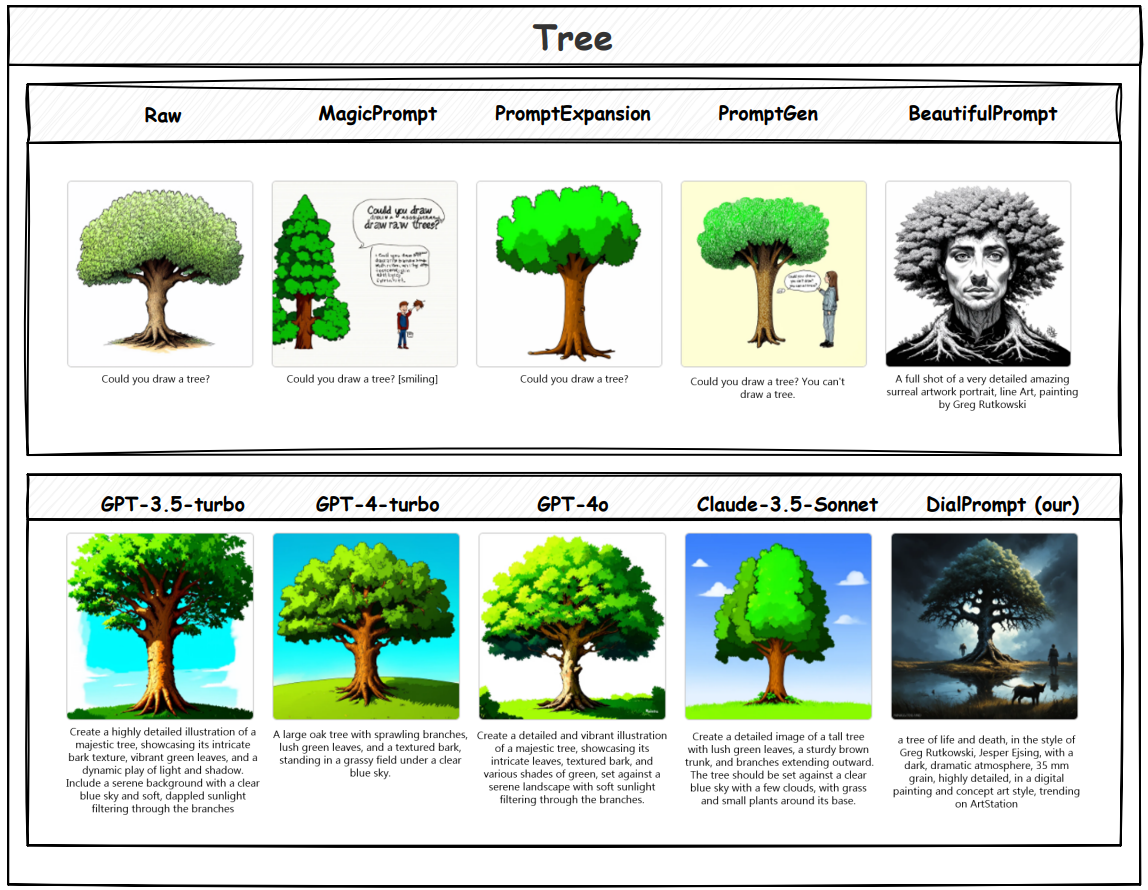}}

\end{figure*}
\begin{figure*}[tp]
 \centering  
 \subfigbottomskip=-2pt 
 \subfigcapskip=-2pt 
 \subfigure{
  \includegraphics[width=0.9\linewidth]{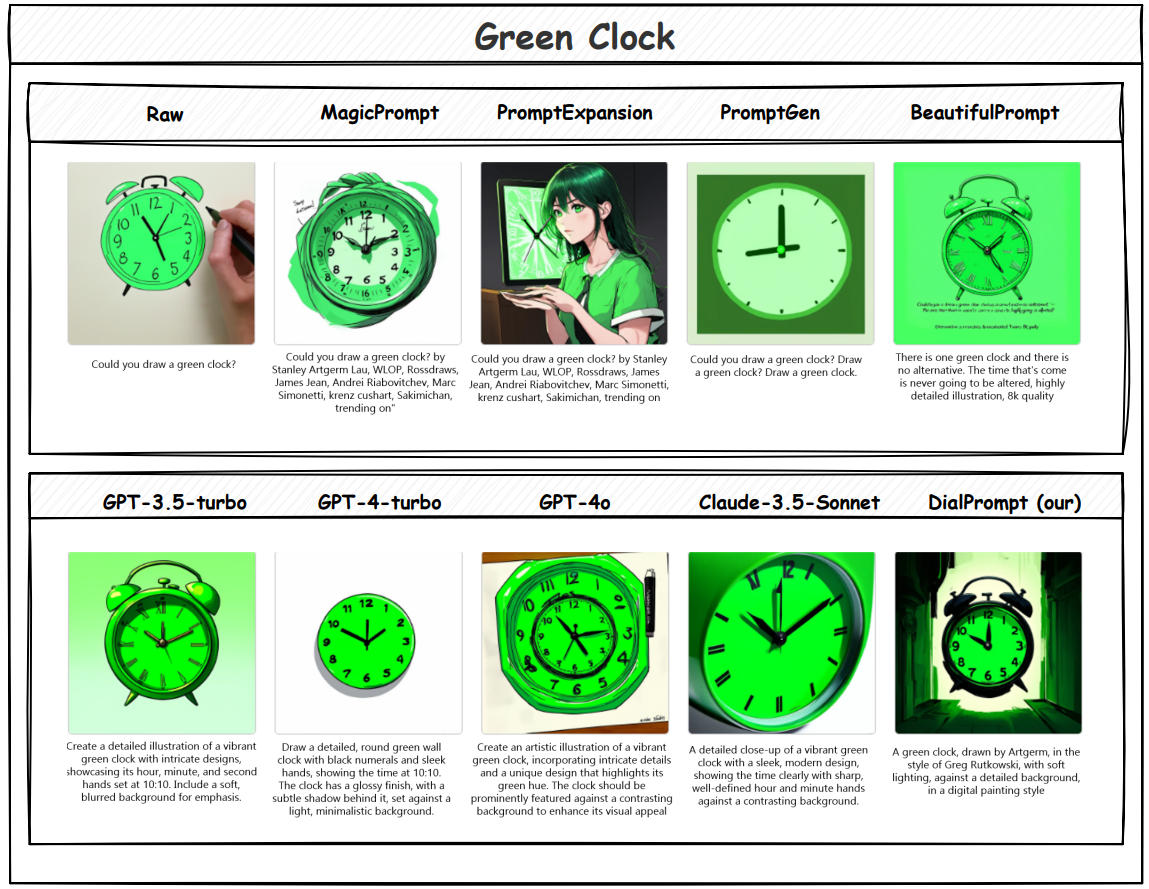}}
   \\
 \subfigure{
  \includegraphics[width=0.9\linewidth]{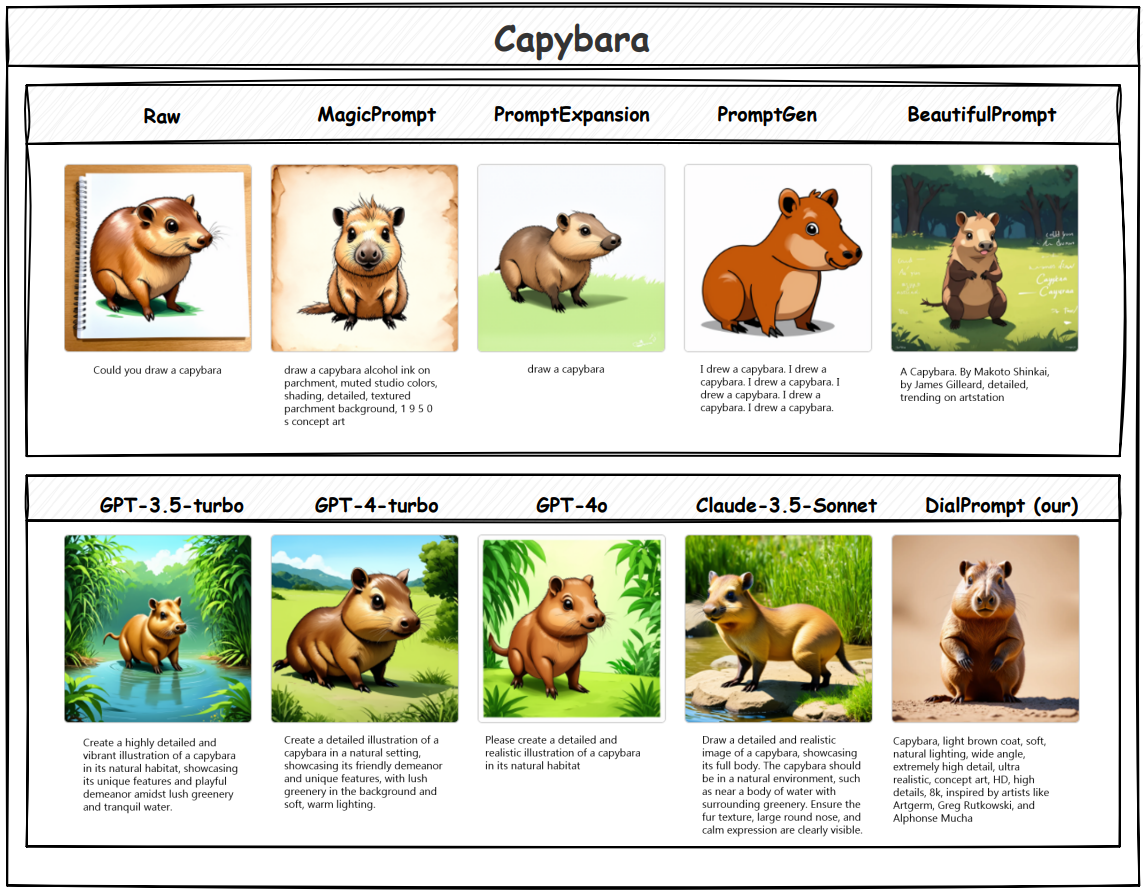}}

\end{figure*}
\begin{figure*}[tp]
 \centering  
 \subfigbottomskip=-2pt 
 \subfigcapskip=-2pt 
 \subfigure{
  \includegraphics[width=0.85\linewidth]{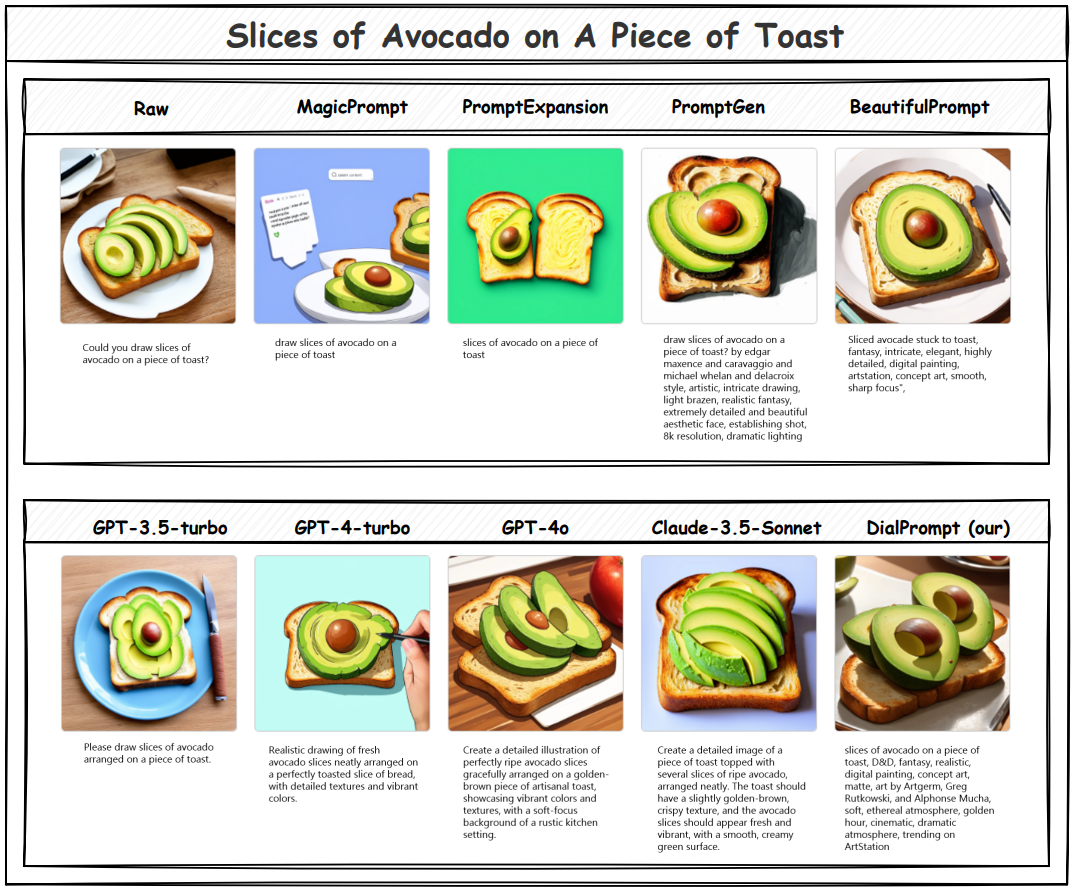}}
   \\
 \subfigure{
  \includegraphics[width=0.85\linewidth]{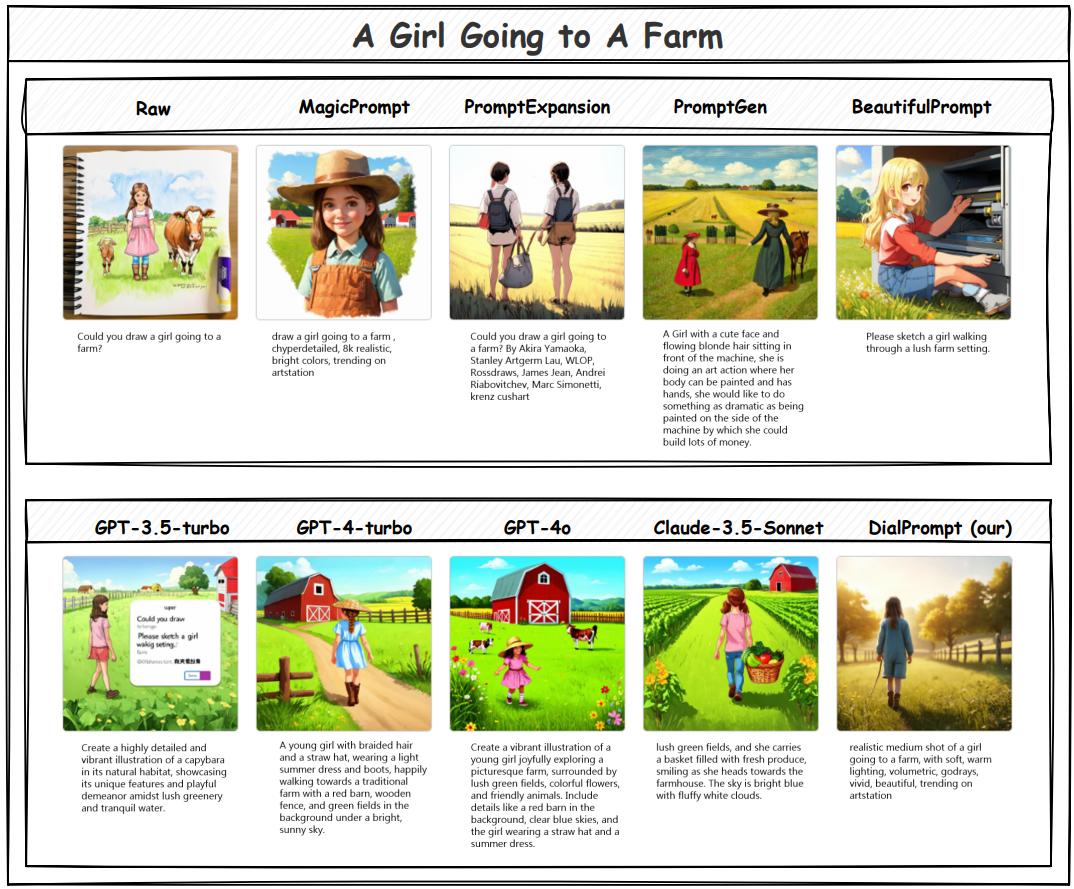}}

\end{figure*}
\begin{figure*}[tp]
 \centering  
 \subfigbottomskip=-2pt 
 \subfigcapskip=-2pt 
 \subfigure{
  \includegraphics[width=0.85\linewidth]{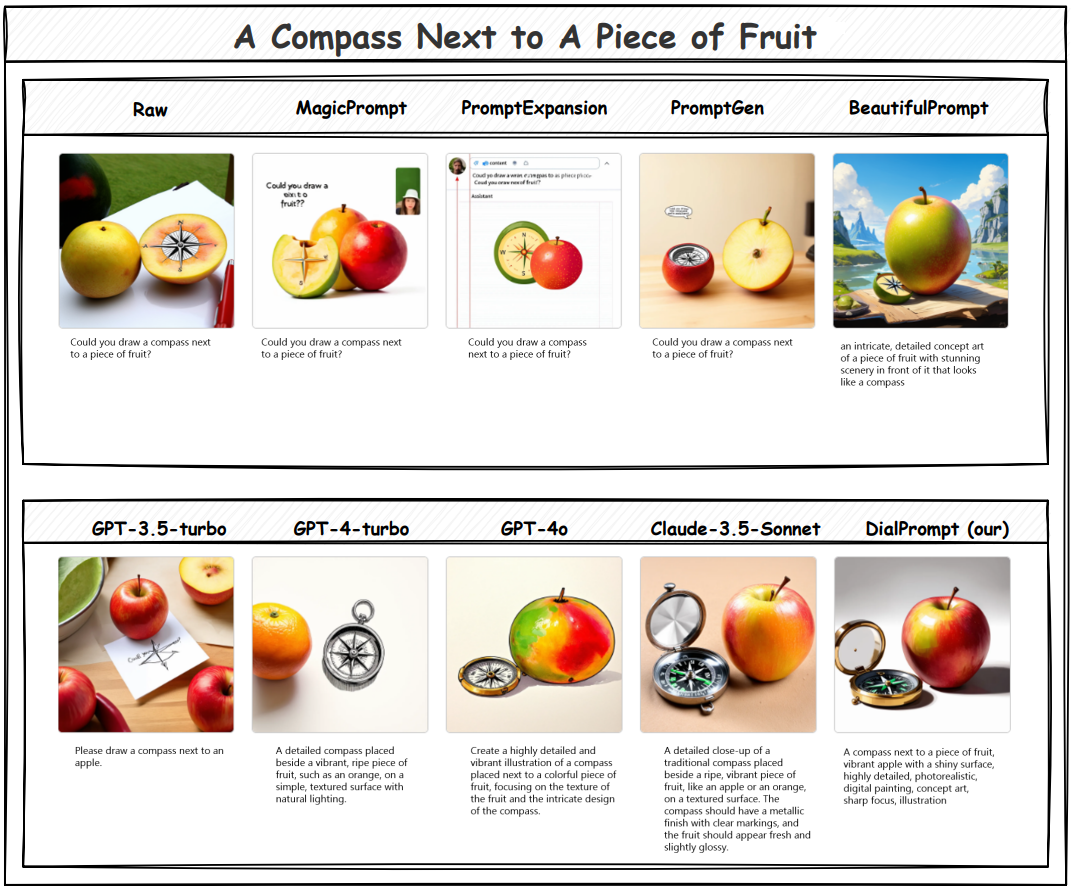}}
   \\
 \subfigure{
  \includegraphics[width=0.85\linewidth]{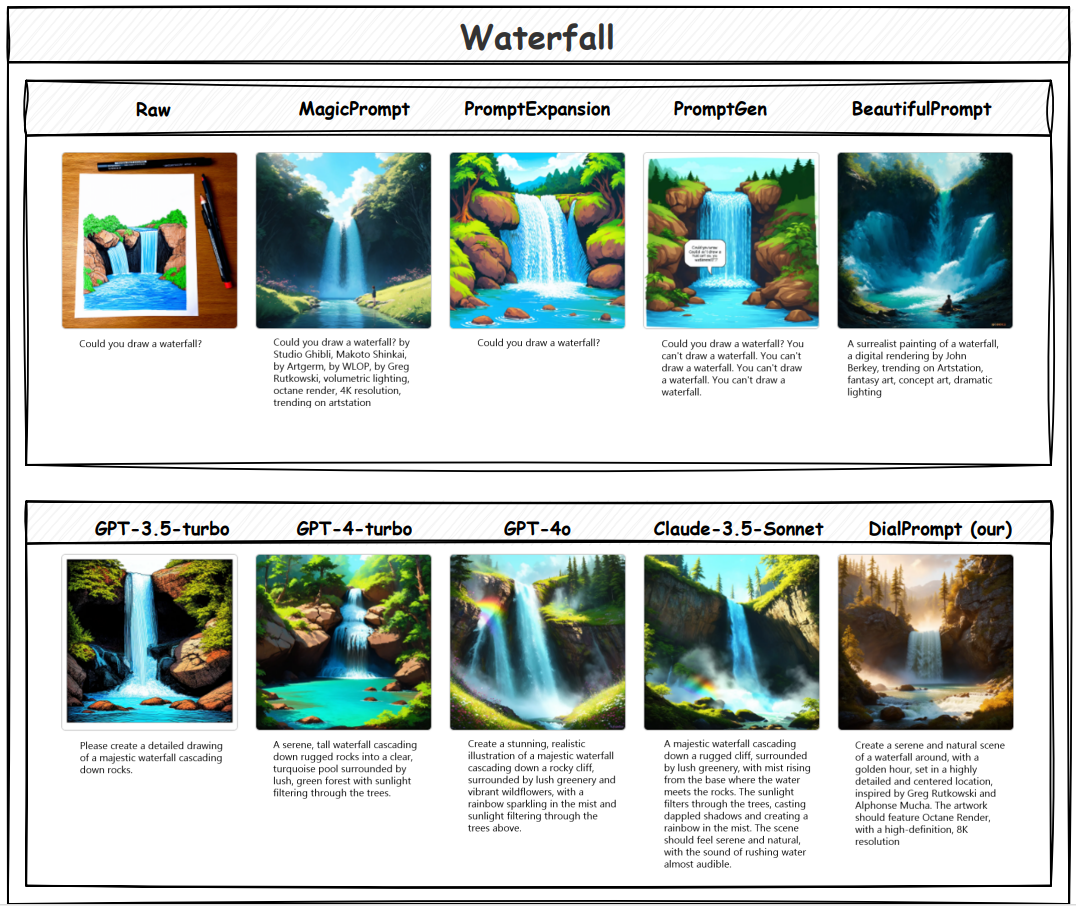}}

\end{figure*}
\end{document}